%% file: LearningMetod_root.tex
\documentclass[a4paper, 10 pt, conference]{ieeeconf}  % Comment this line out if you need a4paper
\usepackage[tmargin = 25mm,lmargin = 20mm, rmargin = 20mm, bmargin = 25mm]{geometry}
\usepackage{mathptmx} % assumes new font selection scheme installed
\usepackage{times} % assumes new font selection scheme installed
\usepackage{amsmath} % assumes amsmath package installed
\usepackage{amssymb}  % assumes amsmath package installed
\usepackage{lipsum}

\usepackage{mathtools, nccmath}
\usepackage{cuted}

\usepackage{multirow}

\usepackage{paralist}
\usepackage{pifont} % For unicode characters with \ding{code} -> Quickreference http://willbenton.com/wb-images/pifont.pdf
\usepackage{tabularx}
\usepackage{booktabs}
\usepackage{color}
\usepackage{colortbl}
\usepackage{color,soul}
\usepackage{float}
\definecolor{schwarz}{rgb}{0.0,0.0,0.0}
\definecolor{black}{rgb}{0.0,0.0,0.0}
\definecolor{darkdarkgray}{rgb}{0.6,0.6,0.6}
\definecolor{darkgray}{rgb}{0.8,0.8,0.8}
\definecolor{lightgray}{rgb}{0.95,0.95,0.95}
\definecolor{white}{rgb}{1.0,1.0,1.0}
\usepackage{stfloats}
\usepackage{blindtext}

\usepackage[export]{adjustbox}

\usepackage{epstopdf} 		 % use epstopdf package to convert eps files into pdf so that they work with pdftex and lualatex. this should give you the feeling that pdftex and lualatex actually support eps.
\usepackage{epstopdf} 		 % use epstopdf package to convert eps files into pdf so that they work with pdftex and lualatex. this should give you the feeling that pdftex and lualatex actually

\usepackage{ifluatex}
\ifluatex
\usepackage{pdftexcmds}
\makeatletter
\let\pdfstrcmp\pdf@strcmp
\let\pdffilemoddate\pdf@filemoddate
\makeatother
\fi
\usepackage{svg}

\usepackage{hhline}
\usepackage{longtable}
\usepackage{amsmath}

\makeatletter
\DeclareFontFamily{U}{tipa}{}
\DeclareFontShape{U}{tipa}{m}{n}{<->tipa10}{}
\newcommand{\arc@char}{{\usefont{U}{tipa}{m}{n}\symbol{62}}}%

\newcommand{\arc}[1]{\mathpalette\arc@arc{#1}}

\newcommand{\arc@arc}[2]{%
	\sbox0{$\m@th#1#2$}%
	\vbox{
		\hbox{\resizebox{\wd0}{\height}{\arc@char}}
		\nointerlineskip
		\box0
	}%
}
\makeatother

\def\@fnsymbol#1{\ensuremath{\ifcase#1\or *\or \dagger\or \ddagger\or
		\mathsection\or \mathparagraph\or \|\or **\or \dagger\dagger
		\or \ddagger\ddagger \else\@ctrerr\fi}}

\newcommand{\ssymbol}[1]{{\@fnsymbol{#1}}}

\usepackage{graphicx}        % standard LaTeX graphics tool
\usepackage{subfig}
\usepackage{caption}

\usepackage[colorlinks=false]{hyperref}
\color{black}

\usepackage{cite}

\usepackage{algcompatible}
\usepackage{algorithm}
\usepackage{algpseudocode}
\algnewcommand{\LeftComment}[1]{\Statex \(\triangleright\) #1}

\usepackage{authblk}
\author[1]{\Large{Francesco Cursi}}
\author[1]{\Large{Guang-Zhong Yang}}
\affil[1]{\large{\textit{Hamlyn Center, Imperial College London}}}
\affil[]{\large{\textit{[f.cursi17,g.z.yang]@imperial.ac.uk}}}

\title{\LARGE \bf
A Robust Regression Approach for Robot Model Learning
}

\IEEEoverridecommandlockouts

\begin{document}

	\maketitle
	\thispagestyle{empty}
	\pagestyle{empty}

	%%%%%%%%%%%%%%%%%%%%%%%%%%%%%%%%%%%%%%%%%%%%%%%%%%%%%%%%%%%%%%%%%%%%%%%%%%%%%%%
	\input{Introduction.tex}

	%%%%%%%%%%%%%%%%%%%%%%%%%%%%%%%%%%%%%%%%%%%%%%%%%%%%%%%%%%%%%%%%%%%%%%%%%%%%%%%
%	\input{Problem.tex}
		%%%%%%%%%%%%%%%%%%%%%%%%%%%%%%%%%%%%%%%%%%%%%%%%%%%%%%%%%%%%%%%%%%%%%%%%%%%%%%%
	\input{Method.tex}

	%%%%%%%%%%%%%%%%%%%%%%%%%%%%%%%%%%%%%%%%%%%%%%%%%%%%%%%%%%%%%%%%%%%%%%%%%%%%%%

	%%%%%%%%%%%%%%%%%%%%%%%%%%%%%%%%%%%%%%%%%%%%%%%%%%%%%%%%%%%%%%%%%%%%%%%%%%%%%%%

	\input{Results.tex}

	%%%%%%%%%%%%%%%%%%%%%%%%%%%%%%%%%%%%%%%%%%%%%%%%%%%%%%%%%%%%%%%%%%%%%%%%%%%%%%%
	\input{Conclusions.tex}
	%%%%%%%%%%%%%%%%%%%%%%%%%%%%%%%%%%%%%%%%%%%%%%%%%%%%%%%%%%%%%%%%%%%%%%%%%%%%%%%

	\addtolength{\textheight}{-12cm}   % This command serves to balance the column lengths

	\section*{COMMENTS}
	This is a preprint of a work which will be published in IROS 2019, Nov 4-8, Macau by the same authors.

	%%%%%%%%%%%%%%%%%%%%%%%%%%%%%%%%%%%%%%%%%%%%%%%%%%%%%%%%%%%%%%%%%%%%%%%%%%%%%%%%
	
	%References are important to the reader; therefore, each citation must be complete and correct. If at all possible, references should be commonly available publications.
	\bibliographystyle{IEEEtran}
			\bibliography{./library}

\end{document}

%% file: Introduction.tex
\begin{abstract}
Machine learning and data
analysis have been used in many robotics fields, especially for modelling. Data
are usually the result of sensor measurements and, as such,
they might be subjected to noise and outliers. The presence
of outliers has a huge impact on modelling the acquired
data, resulting in inappropriate models. In this work a novel
approach for outlier detection and rejection for input/output
mapping in regression problems is presented. The robustness
of the method is shown both through simulated data for linear
and nonlinear regression, and real sensory data. Despite being
validated by using artificial neural networks, the method can
be generalized to any other regression method.
\end{abstract}
\section{Introduction}
Machine learning is defined  as a set of methods that can
automatically detect patterns in data, and then use the uncovered patterns to predict future
data  and perform decision making under uncertainty \cite{Murphy2012}.
In the field of robotics, machine learning has been widely used to accurately approximate models of robots, without the need of analytical models, which may be hard to obtain due to the complexity of the system \cite{Nguyen-Tuong2011}.
Many algorithms have been developed for solving the regression problem in robot modelling \cite{Sigaud2011}, however, in order to build good models, data must be carefully analysed, since outliers and noise may affect the results. Most of the proposed methods do not address the problem of outlier rejection and usually assume that data points follow a known distribution (typically Gaussian).  \\

An outlier is defined as a data point significantly different from the others \cite{Hawkins1980} and the presence of unwanted data points may lead to a wrong model describing the relationship between the input and the output values \cite{Rousseeuw2003}. 
Different approaches exist to detect outliers in datasets \cite{Aggarwal2013}, yet, all these outlier detection method either rely on only one particular method, or on the knowledge of the data statistical distribution. Moreover, they can be regarded to as a sort of preprocessing approaches. As a matter of fact, first outliers have to be detected with one of these methods, and, only afterwards, regression methods can be applied to the good data points to find the appropriate model. Robust regression approaches such as iteratively reweighed least squares \cite{Street1988} or random sample consensus (RANSAC) \cite{Fischler1981} for linear regression, instead, allow to neglect outliers while the regression process takes place.

In this work a novel approach for robust data modelling and input/output mapping is presented. The method doesn't require any preprocessing to identify outliers, since they are automatically found while learning the model. Moreover, no assumption on the data distribution is required.  The proposed method has been validated by using neural networks for regression, yet it can be generalized to any other regression method such as linear regression, Gaussian process regression, etc.\\

The paper is thus structured as follows.

Section \ref{sec:Method} presents the proposed method. Section \ref{sec:Results} shows the results on simulated and real data. For the simulated data, two cases are analyzed: linear and nonlinear regression. For the real application, the method is applied to model the dynamics of a tendon-driven surgical robot. Conclusions are then drawn in Section \ref{sec:Conclusions}.

%% file: Method.tex
\section{Method}\label{sec:Method}
Given a dataset of input points $\mathbf{x} \in \mathbb{R}^{n_{in}}$ and output points $\mathbf{y} \in \mathbb{R}^{n_{out}}$, the goal of regression is to find best the relationship between the two, meaning
\begin{equation}
	\mathbf{y \sim f(x) }\ 
\end{equation}
where $\mathbf{f}(\cdot)$ can be any linear or nonlinear function. Artificial neural network (ANN) can model
any suitably smooth function, given enough hidden units, to any desired level of accuracy
\cite{Hornik1991}. They are thus capable of representing complicated behaviours, without the need of knowing any mathematical or physical model.
Nevertheless, it has been shown that NN behaviour is influenced by outliers \cite{Khamis2005,Liano1996}.

In order to build a model which is not affected by bad data, outliers must be detected and somehow neglected. Given $n$ data points their estimated output value for each output component is $\tilde{y}_{i,j}$ for $i = 1...n_{out}$, $j = 1...n $.
For each output dimension the vector of residuals is computed $r_i =\begin{bmatrix}y_{i,1}-\tilde{y}_{i,1}\ , & . &. &. &\ ,y_{i,n}-\tilde{y}_{i,n} \end{bmatrix}$. The median $m_i$ and the median absolute deviation $MAD_i$ of each residual vector are calculated, and the threshold is then set to $t_i = \gamma*MAD_i$, with $\gamma$ being a positive constant.

Once the medians and the thresholds have been retrieved, each data sample is assigned an output weight as follows:
\begin{equation}
 v = \frac{r_{i.j}-m_i}{t_i}\ ,
 W_{i,j} = e^{-7v^8}\ .
\end{equation}
A data point for a certain output component is thus an outlier if its residual is too far from the median of the residuals.

In order to build a robust model from a given dataset, an iterative re-weighing process is performed. At first, each sample for each output component is assigned a unitary weight and a first model is built. Then the inliers and their weights are computed, and the model is refined with the new weights. The process continues until a desired number of refinements is reached.

%% file: Results.tex
\section{Results}\label{sec:Results}
In this Section the proposed method is tested on simulated data and real data. For the simulation, it is applied to linear and non linear regression. The method is compared to traditional ANN, RANSAC for the linear regression, and Gaussian Process Regression for the nonlinear regression. For the real experiment, the robust method is applied to model the dynamics of a tendon-driven robot.
\begin{figure*}
	\centering
	\includegraphics[width = \textwidth]{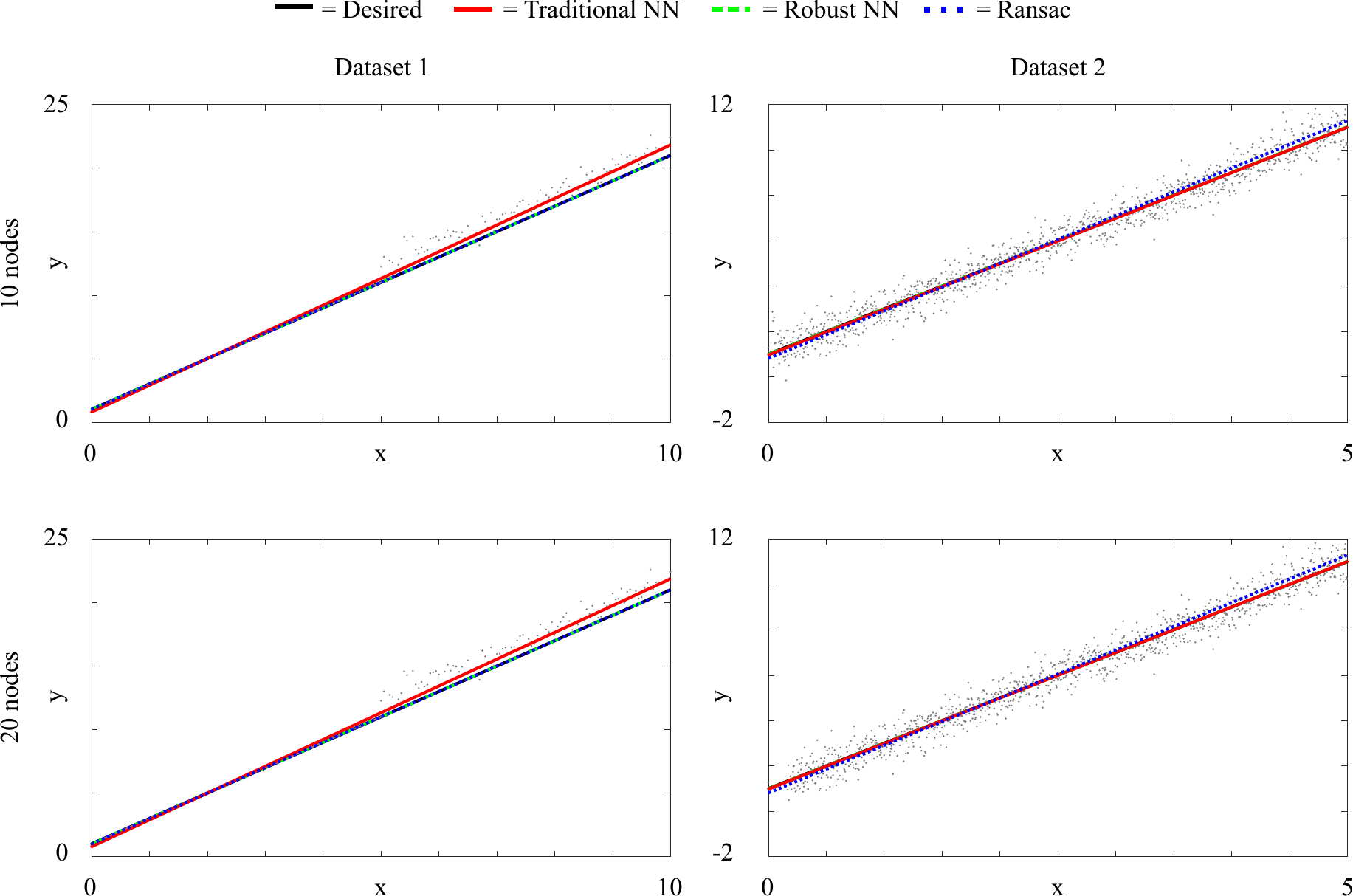}
	\caption{Comparison of the results for the linear regression on the two datasets.}
	\label{fig:LinRes}
\end{figure*}

\begin{figure*}
	\centering
	\includegraphics[width = \textwidth]{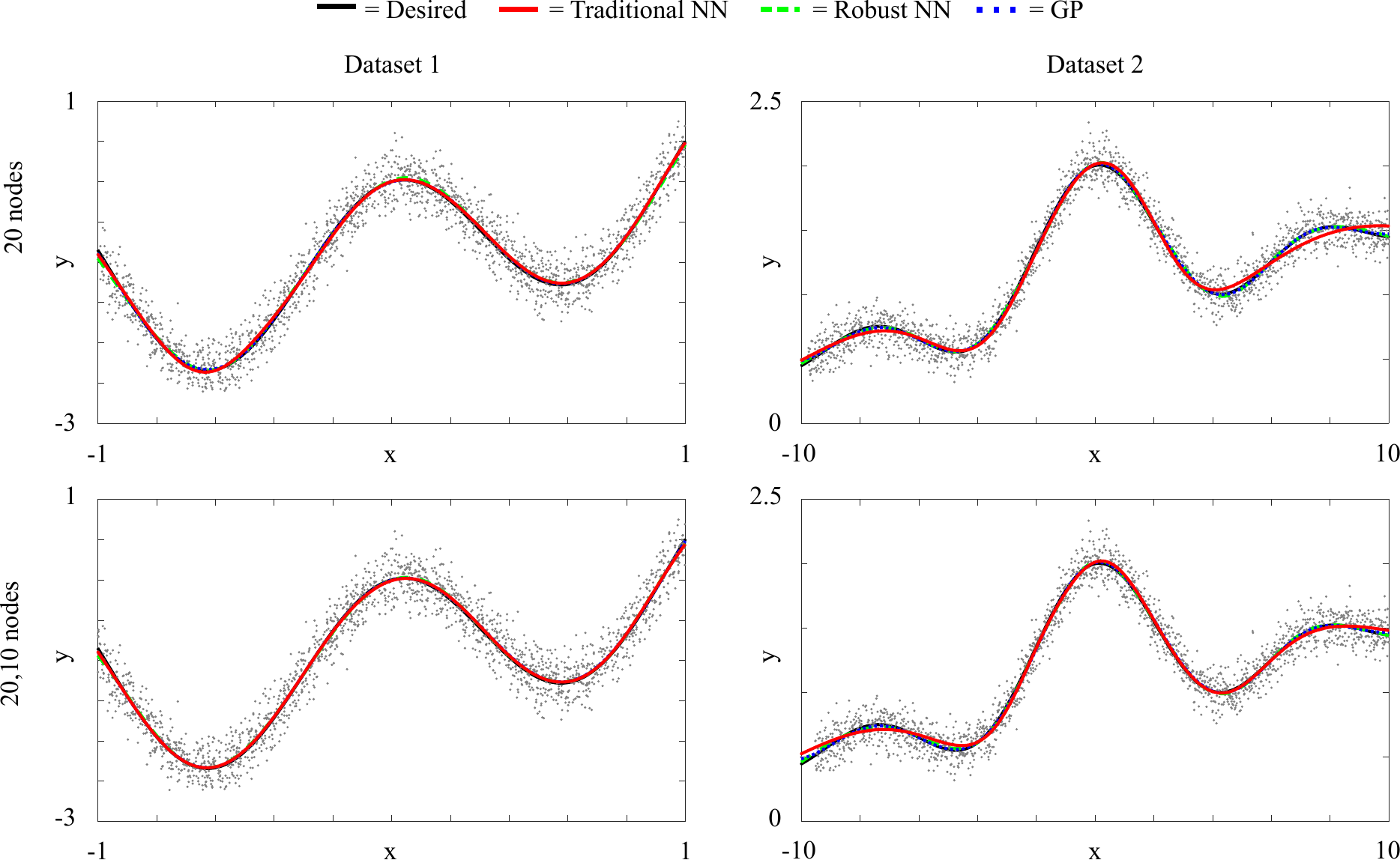}
	\caption{Comparison of the results for the nonlinear regression on the two datasets without outliers, but only Gaussian noise.}
	\label{fig:NonLinRes_NoOut}
\end{figure*}

\begin{figure*}
	\centering
	\includegraphics[width = \textwidth]{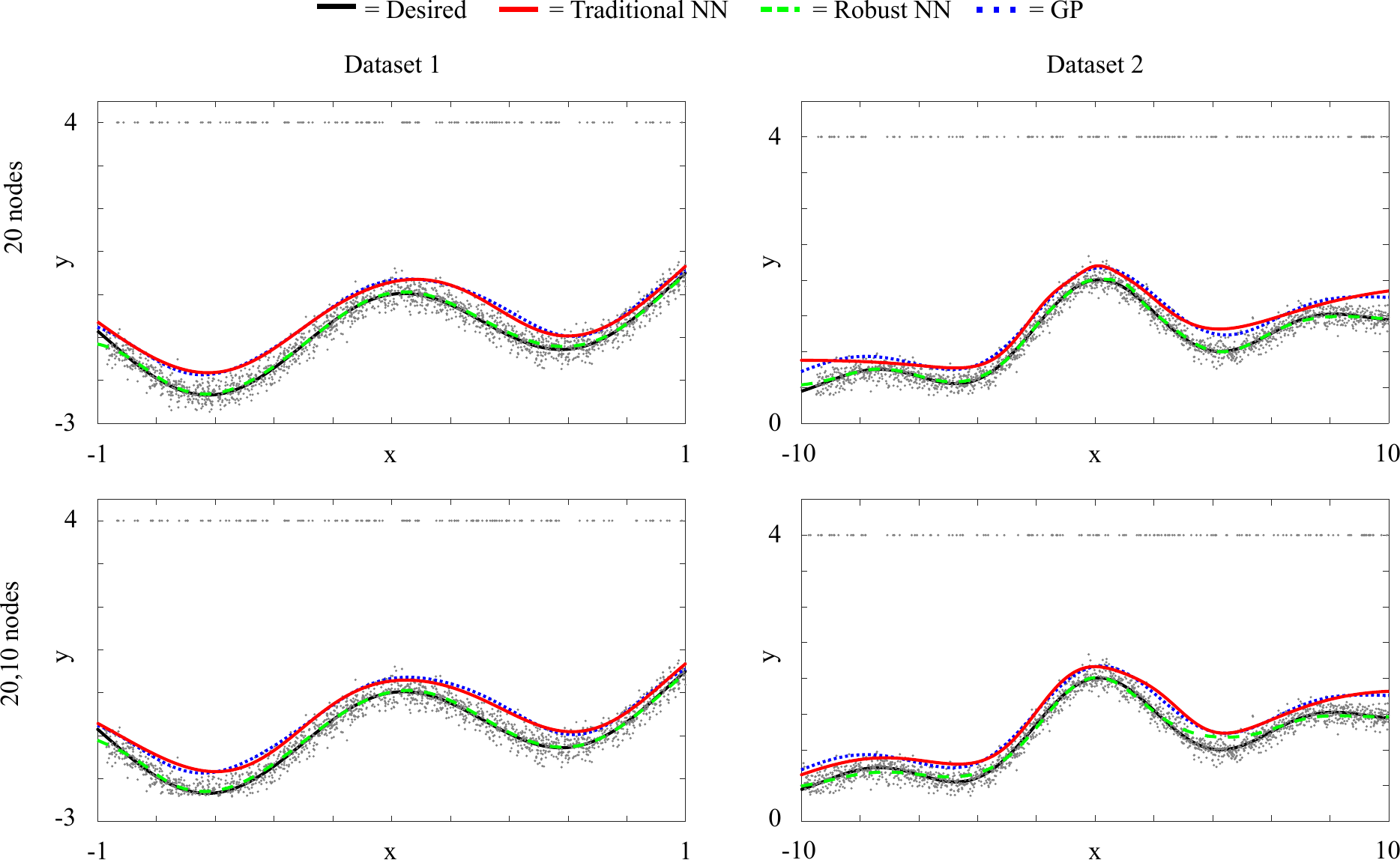}
	\caption{Comparison of the results for the nonlinear regression on the two datasets with Gaussian noise and outliers.}
	\label{fig:NonLinRes_Out}
\end{figure*}

\begin{figure*}
	\centering
	\includegraphics[width = \textwidth]{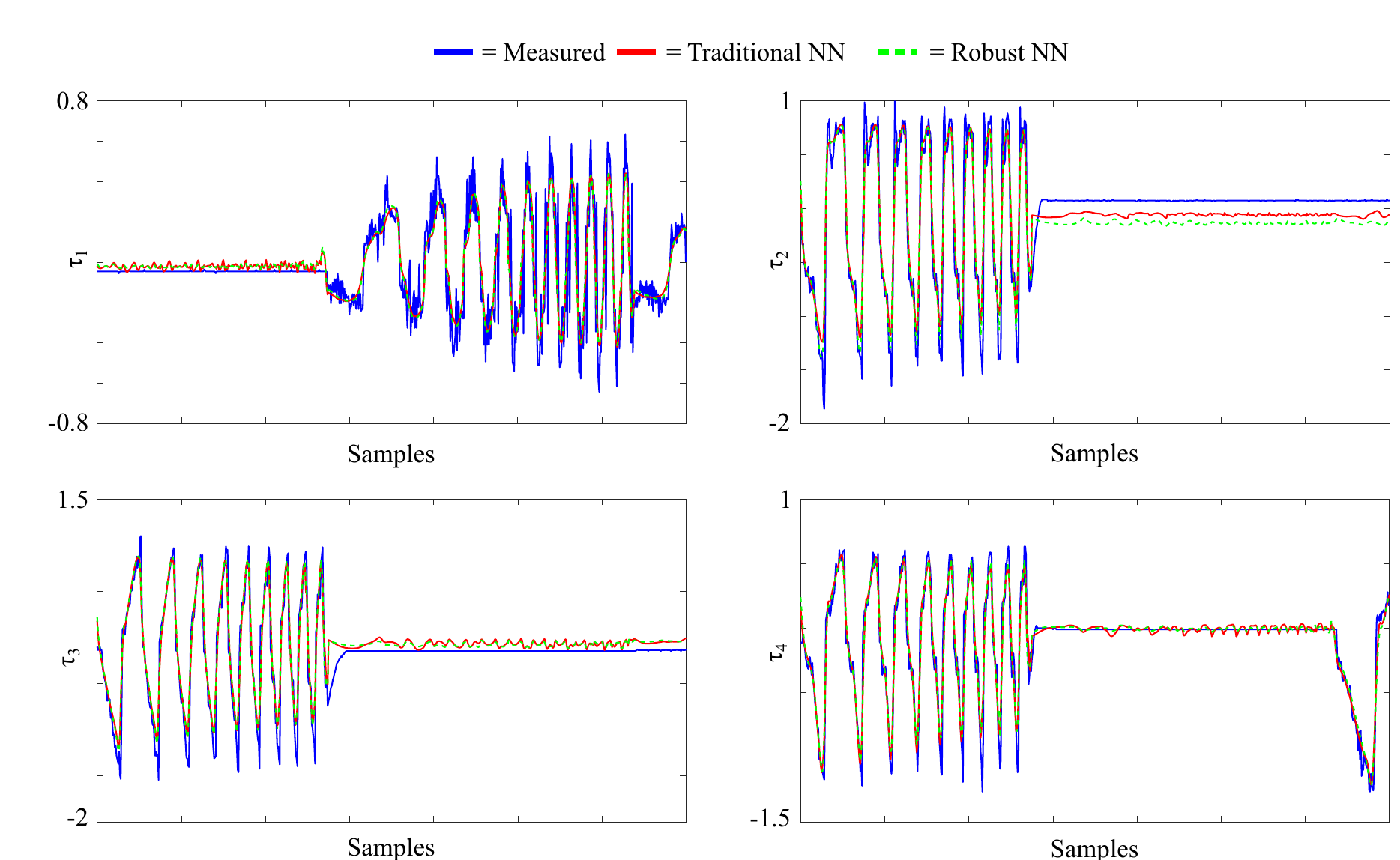}
	\caption{Example of the results for modelling the robot dynamics. Here a small window of the whole dataset is shown. All measures are in mNm}
	\label{fig:uIges}
\end{figure*}

\subsection{Simulation Data}
	\begin{table}[h]
    \centering
    \begin{tabular}{|c|c|c|c|}
    \hline
    \multicolumn{4}{|c|}{\textbf{Linear Dataset 1}}\\
    \hline
    & &$\mathbf{R^2}$ & \textbf{RMSE} \\
    \hline
    \multirow{2}{*}{\textbf{Robust NN}} &10 nodes &0.9927 &0 \\
    &20 nodes &0.9927 &0 \\
    \hline
    \multirow{2}{*}{\textbf{Traditional NN}} &10 nodes &0.9957 &0.2337 \\ 
    &20 nodes &0.9957 &0.2457 \\
    \hline
    \textbf{RANSAC} & &0.9927 &0 \\
\hline
\hline
%%%%%%%%%%%%%%%%%%%%%%%%%%%%%%%%%%%%%%%%%%%%%%%%%%%%%%%%%%%%%%%%
    \multicolumn{4}{|c|}{\textbf{Linear Dataset 2}}\\
    \hline
    & &$\mathbf{R^2}$ & \textbf{RMSE} \\
    \hline
    \multirow{2}{*}{\textbf{Robust NN}} &10 nodes &0.9706 &0.0105 \\
    &20 nodes &0.9706 &0.0076 \\
    \hline
    \multirow{2}{*}{\textbf{Traditional NN}} &10 nodes &0.9706 &0.0199 \\ 
    &20 nodes &0.9706 &0.0114 \\
    \hline
    \textbf{RANSAC} & &0.9679 &0.1498 \\
\hline
    \end{tabular}
    \caption{$R^2$ and RMSE between the computed models and the desired mapping for the linear case.}
    \label{tab:Lin}
\end{table}
For the linear regression example, two different datasets are used. In both cases the desired function is described by $y = 2x+1$.
In the first dataset, some noisy data is added only in  a certain input region. The noisy data is generated from a random Gaussian distribution with mean ($\mu$) equal to 1 and standard deviation ($\sigma$) of 0.5. In total 1100 data points are used, with 100 points being noisy.
In the second dataset, instead, all the 1100 points are corrupted by noise (with $\mu = 0,\ \sigma = 0.5$). For building the robust model, the threshold has been set to $t = 2MAD$ and 5 refinements are executed.
Figure \ref{fig:LinRes} shows the results comparing the robust proposed method, the traditional ANN, and RANSAC method \cite{Fischler1981}.

For both the robust NN and traditional NN, the datset is divided randomly into train set ($80\%, 10\%,\ 10\% $). Different network architectures are used: on single hidden layer with 10 nodes and with 20 nodes. In both cases, linear activation functions for each node are used. For RANSAC, 2 samples are used, the maximum distance is set equal to $\sigma$, and the distance function being the euclidean distance between the output and the expected output value. Table \ref{tab:Lin} shows the goodness of fit expressed in terms of the $R^2$ \cite{Allen1997} and the RMSE between the computed models and the desired one.

In the first datsets, RANSAC and the robust method perform very close to each other finding the correct mapping. Traditional NN, instead, doesn't mange to obtain good results, with the model being biased by the noisy data. In the second dataset, instead, RANSAC is the one the performs worst. The proposed robust method, instead, performs the best.\\

For the nonlinear case two different mappings are sought. In the first mapping the desired function is described by $y = -0.5x^3+\cos(5x)+e^x-2$ and the Gaussian noise by $\mu = 0,\ \sigma = 0.2$, whereas in the second the function is $y = 1+0.05x+\frac{\sin(x)}{x}$ and the noise by $\mu = 0,\ \sigma = 0.1$. In both cases 2000 data points are used. Two different network architectures are used: one single hidden layer with 20 nodes and a two-hidden-layer structure with 20 and 10 nodes each. The robust and traditional ANN methods are here compared to Gaussian Process Regression (GPR). Figure \ref{fig:NonLinRes_NoOut} shows the results. The same threshold and refinements of the linear case are used. In order to see the behaviour when points outside a Guassian distribution are present, 150 outliers are also added (Figure \ref{fig:NonLinRes_Out}). Table \ref{tab:NonLin} reports the error metrics for the different methods on the different datasets.

When no outliers are present, all methods perform pretty well and the models are pretty close one to each other. On the second datasets, however, traditional NN is the one that performs the worst. When outliers are included, instead, the models from GPR and traditional NN are compromised. The two methods perform very similarly, with large errors. The robust approach, conversely, manages to keep errors small, even if a bit higher than in the case with only Gaussian noise.  

%%%%%%%%Nonlinear Table
	\begin{table*}[h]
    \centering
    \begin{tabular}{|c|c|c|c|c|c|}
    \hline
    \multicolumn{6}{|c|}{\textbf{Nonlinear Dataset 1}} \\
    \hline
    & &\multicolumn{2}{|c|}{\textbf{No Outliers}} & \multicolumn{2}{|c|}{\textbf{ With Outliers}}\\
    \hline
    & &$\mathbf{R^2}$ & \textbf{RMSE} &$\mathbf{R^2}$ & \textbf{RMSE} \\
    \hline
    \multirow{2}{*}{\textbf{Robust NN}} &10 nodes &0.9361 &0.0276 &0.1722 &0.0521 \\
    &20,10 nodes &0.9364 &0.0197 &0.1719 &0.0487 \\
    \hline
    \multirow{2}{*}{\textbf{Traditional NN}} &10 nodes &0.9365 &0.0215 &0.2302 &0.3825 \\ 
    &20,10 nodes &0.9364 &0.0152 &0.2292 &0.3784 \\
    \hline
    \textbf{GPR} & &0.9364 &0.0143& 0.2320 &0.3963 \\
\hline
\hline
%%%%%%%%%%%%%%%%%%%%%%%%%%%%%%%%%%%%%%%%%%%%%%%%%%%%%%%%%%%%%%%%
        \multicolumn{6}{|c|}{\textbf{Nonlinear Dataset 2}} \\
    \hline
    & &\multicolumn{2}{|c|}{\textbf{No Outliers}} & \multicolumn{2}{|c|}{\textbf{ With Outliers}}\\
    \hline
    & &$\mathbf{R^2}$ & \textbf{RMSE} &$\mathbf{R^2}$ & \textbf{RMSE} \\
    \hline
    \multirow{2}{*}{\textbf{Robust NN}} &10 nodes
    &0.9549 &0.0123 &0.2165 &0.0632 \\
    &20,10 nodes &0.9549 &0.0109 &0.2172 &0.0212 \\
    \hline
    \multirow{2}{*}{\textbf{Traditional NN}} &10 nodes 
    &0.9510 &0.0309 &0.2720 &0.2147 \\ 
    &20,10 nodes &0.9535 &0.0238 &0.2699 &0.2196 \\
    \hline
    \textbf{GPR} & &0.9551 &0.0102 &0.2728 &0.2120 \\
\hline
    \end{tabular}
    \caption{$R^2$ and RMSE between the computed models and the desired mapping for the nonlinear cases.}
    \label{tab:NonLin}
\end{table*}
%%%%%%%%%%%%%%%%%%%%%%%%%%%%%%%%%%

\subsection{Robot Dynamic Modelling}
As an example of real life application, we were interested in collecting data for learning the inverse dynamics and nolinearities of the Micro-IGES robotic surgical tool \cite{Shang2017}. This robot is tendon-driven and the major causes of nonlinearities are due to the routing and elasticity of the tendons, and friction in the joints and along the tendons. Because of the motor to joint mapping,the inverse dynamics of the system can be expressed in therms of the motor values $\mathbf{\theta}$, $\mathbf{\dot{\theta}}$, $\mathbf{\ddot{\theta}}$, and torques $\mathbf{\tau}$ as:
\begin{equation}
\mathbf{\tau} = \Gamma(\mathbf{\theta,\dot{\theta},\ddot{\theta}})\ .
\end{equation}
Only 4-dof of the articulated part of the robot are considered and, given the motor values ($\mathbf{\theta,\dot{\theta},\ddot{\theta}} \in \mathbb{R}^4$) as inputs, the corresponding $\mathbf{\tau} \in \mathbb{R}^4$ is sought. 

For the learning the model a single neural network with two hidden layers of 20 and 10 neurons was used. Each motor was excited with a sinusoidal wave of linearly increasing frequency within the range $[0,1]$ Hz. Figure \ref{fig:uIges} shows the results by using the traditional and robust approaches on a subset of the data and Table \ref{tab:uIgeslearning} reports the $R^2$ and the RMSE between the two models and the measured values in the whole dataset.

Both methods perform well, with good $R^2$ and small errors. However, it can be noted that the robust method allows to have smoother mapping, being less influenced by unwanted data.

	\begin{table}[h]
    \centering
    \begin{tabular}{|c|c|c|c|c|c|c|}
    \hline
    \multicolumn{7}{|c|}{\textbf{Robust NN Method}} \\
    \hline
         &  \multicolumn{3}{|c|}{$\mathbf{R^2}$} & \multicolumn{3}{|c|}{$\mathbf{RMSE}\ (mNm)$}\\
         \hline
         \textbf{i} & \textbf{train} & \textbf{test} & \textbf{validation} & \textbf{train} & \textbf{test} & \textbf{validation} \\
         \hline
         1 & 0.939&  0.945&	0.941& 0.0454 &	0.0439  & 0.0445 \\
         \hline
         
                  2 & 0.939 &  0.939&	0.943 & 0.1326 &0.1324&	0.1293 	\\
         \hline
         
                  3 & 0.935&  0.935&	0.935 & 0.1160	& 0.1190& 0.1155	 \\
         \hline
         
                  4 & 0.969&  0.968&	0.969 & 0.0649 &0.0656&	0.0643	  	\\
         \hline
         \hline
         
         %Traditional
           \multicolumn{7}{|c|}{\textbf{Traditional NN Method}}\\
           \hline
           &  \multicolumn{3}{|c|}{$\mathbf{R^2}$} & \multicolumn{3}{|c|}{$\mathbf{RMSE}\ (mNm)$}\\
         \hline
         \textbf{i} & \textbf{train} & \textbf{test} & \textbf{validation} & \textbf{train} & \textbf{test} & \textbf{validation} \\
         \hline
         1 & 0.947  &0.951  &	0.946 & 0.0426&	0.0417 & 0.0424 		 \\
         \hline
         
                  2 & 0.948 &  0.946&	0.952 & 0.1219 &0.1243&	0.1192 	\\
         \hline
         
                  3 & 0.941&  0.941&	0.942 & 0.1108	& 0.1132& 0.1094	 \\
         \hline
         
                  4 & 0.972&  0.972&	0.972 & 0.0609 &0.0618&	0.0609	  	\\
         \hline
         \hline
    \end{tabular}
    \caption{$R^2$ and RMSE between the learnt torque models and the collected data for the Micro-IGES robot on the different datasets.}
    \label{tab:uIgeslearning}
\end{table}

%% file: Conclusions.tex
\section{Conclusions}\label{sec:Conclusions}
In conclusion, a novel learning algorithm has been presented in this work. The algorithm doesn't make any assumption on the data distribution and allows to have outliers identification while performing the regression, without the need of any preprocessing. Results show that the method is able to neglect undesired data points and, in turn, to produce robust models, minimally influenced by outliers. However, higher robustness comes at a price of higher computational efforts. In all tests the robust method took more time than traditional NN to build the model. In the nonlinear case, the computational time was comparable to that of GPR.

In the next future, the algorithm will be improved in order to make it robuster and reduce the computational time, so that to apply it to online learning.  